# Gaussian Material Synthesis

KÁROLY ZSOLNAI-FEHÉR, TU Wien
PETER WONKA, KAUST
MICHAEL WIMMER, TU Wien

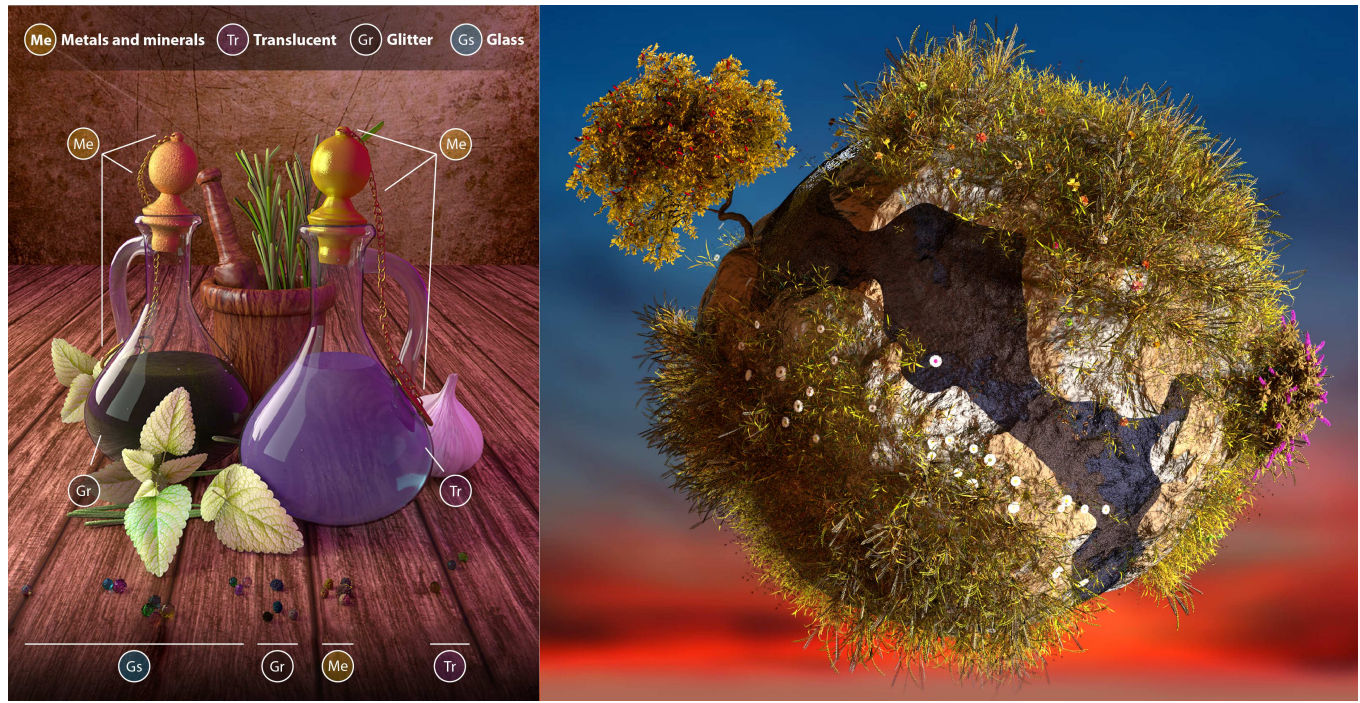

Fig. 1. Our system opens up the possibility of rapid mass-scale material synthesis for novice and expert users alike. This method takes a set of user preferences as an input and recommends relevant new materials from the learned distributions. On the left, we populated a scene with metals and minerals, translucent, glittery and glassy materials, each of which was learned and synthesized via our proposed technique. The image on the right showcases rich material variations for more than a hundred synthesized materials and objects for the vegetation of the planet. The learning and recommendation steps take less than a minute.

We present a learning-based system for rapid mass-scale material synthesis that is useful for novice and expert users alike. The user preferences are learned via Gaussian Process Regression and can be easily sampled for new recommendations. Typically, each recommendation takes 40-60 seconds to render with global illumination, which makes this process impracticable for real-world workflows. Our neural network eliminates this bottleneck by providing high-quality image predictions in real time, after which it is possible to pick the desired materials from a gallery and assign them to a scene in an intuitive manner. Workflow timings against Disney's "principled" shader reveal that our system scales well with the number of sought materials, thus empowering even novice users to generate hundreds of high-quality material models without any expertise in material modeling. Similarly, expert users experience a significant decrease in the total modeling time when populating a scene with materials. Furthermore, our proposed solution also offers controllable recommendations and a novel latent space variant generation step to enable the real-time fine-tuning of materials without requiring any domain expertise.

## 1 INTRODUCTION

Light transport simulations are the industry standard way to create high-quality photorealistic imagery. This class of techniques

Authors' addresses: Károly Zsolnai-Fehér, TU Wien, Favoritenstrasse 9-11/193-02, Vienna, Austria, 1040, zsolnai@cg.tuwien.ac.at; Peter Wonka, KAUST, Al Khwarizmi Bldg 1, Thuwal, Kingdom of Saudi Arabia, 23955-6900, pwonka@gmail.com; Michael Wimmer, TU Wien, Favoritenstrasse 9-11/193-02, Vienna, Austria, 1040, wimmer@cg.tuwien.ac.at.

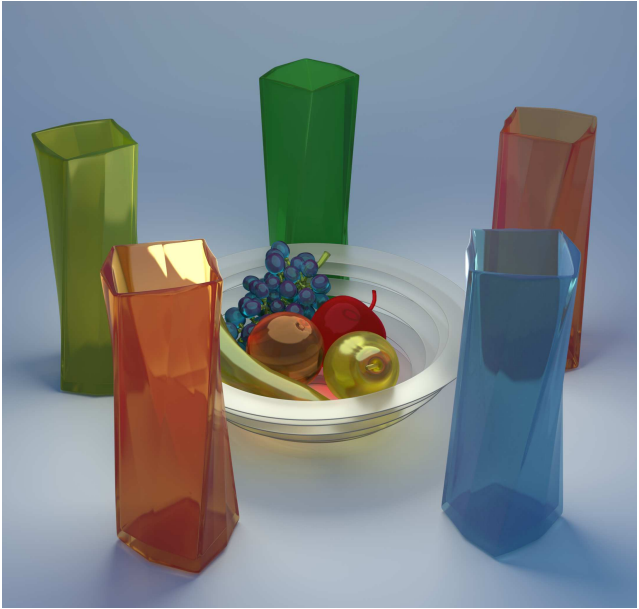

Fig. 2. Glassy materials learned and synthesized by our technique using 150 training samples, 46 of which obtained a score greater than zero.

enjoys a variety of use in architectural visualization, computer animation, and is rapidly becoming the choice of many mass media and entertainment companies to create their feature-length films. Beyond using physically accurate algorithms, the presence of complex material models and high-resolution geometry are also important factors in creating convincing imagery. Choosing the perfect material models is a labor-intensive process where an artist has to resort to trial and error, where each try is followed by the lengthy process of rendering a new image. In this work, we focus on providing tools for rapid mass-scale material synthesis to ease this process for novice and expert users alike – Figures 2 and 3 showcase example scenes where our technique was used to learn glassy and translucent materials. Instead of using a set of specialized shaders for each desired material class, it is generally possible to design one expressive shader that can represent a large swath of possible material models at the cost of increased complexity, which is often referred to as a "principled" or "uber" shader in the rendering community. Each parameterization of such a shader corresponds to one material model. Our strategy is to create a principled shader that is highly expressive, where the complexity downside is alleviated by the fact that the user never has to directly interact with it. To achieve this, we harness the power of three learning algorithms and show that this approach has several advantages compared to the classical workflow (i.e., direct interaction with a "principled" shader): when using our framework, the user is presented with a gallery where scores can be assigned to a set of proposed material models. These scores are used as training samples to adapt to these preferences and create new material recommendations. These recommendations are *controllable*, i.e., the user can choose the amount of desired variety in the output distribution, and our system retains the degree of physical correctness of its underlying shader. Normally, each of these new recommendations would have to be rendered via global illumination, leading to long waiting times. To alleviate this, we have replaced the renderer with a neural network that is able to predict these images in real time. We use a third learning algorithm to perform variant generation, which enables the user to fine-tune previously recommended materials to their liking in real time without requiring any domain expertise.

Furthermore, we explore combinations of these learning algorithms that offer useful real-time previews and color coding schemes to guide the user's attention to regions that are ample in variants with a high expected score and are also deemed *similar* to the fine-tuned input. We also show that our framework scales well with the number of sought materials and that it offers favorable modeling times compared to the classical workflow.

In summary, we present the following contributions:

- a framework for mass-scale material learning and recommendation that works with any high-dimensional principled shader,
- a Convolutional Neural Network to enable the visualization of the recommended materials in real time,
- a latent space variant generation technique that helps the user to intuitively fine-tune the recommended materials in real time,
- a novel way to combine all three learning algorithms to provide color coding for efficient latent-space exploration and real-time previews.

We provide our pre-trained neural network and the source code for the entirety of this project.

## 2 PREVIOUS WORK

*Material modeling.* Many material modeling workflows start with an acquisition step where a real-world material is to be measured with strobes and turntables [Miyashita et al. 2016], screens and cameras [Aittala et al. 2013] or other equipment to obtain a digital version of it that mimics its reflectance properties. Most learning-based techniques focus on reconstructing real-life material models with an SVBRDF[1] model from a flash and one no-flash image pair [Aittala et al. 2015], or remarkably, even from one input photo [Aittala et al. 2016]. In this work, we focus on a different direction where no physical access to the sought materials or additional equipment is required. Previous database-driven methods contain acquired data for a vast number of possible materials to populate a scene, however, they are typically very sizable and either cannot produce new materials on the fly [Bell et al. 2013], or are lacking in more sophisticated material representations (e.g., BSSRDFs[2]) [Matusik 2003].

A different class of methods focuses on directly editing BRDF models within a scene [Ben-Artzi et al. 2006; Cheslack-Postava et al. 2008; Sun et al. 2007], where a typical use-case includes a user populating a scene with one material at a time. Recent studies have also shown that designing intuitive control interfaces for material manipulation is non-trivial [Kerr and Pellacini 2010; Serrano et al. 2016]. To alleviate this, our framework does not expose any BSDF parameters to the user, but learns the user preferences directly and is able to rapidly recommend desirable material models on a mass scale. Chen et al.'s work [2015] operates similarly by populating a scene with materials based on an input color scheme. We expect that this, along with other recent color mixing methods (e.g., Shugrina et al.'s work [2017]) can be combined with our proposed technique to enhance the process of assigning a collection of materials to a scene.

---
[1]Spatially Varying Bidirectional Reflectance Distribution Function
[2]Bidirectional Subsurface Scattering Reflectance Distribution Function

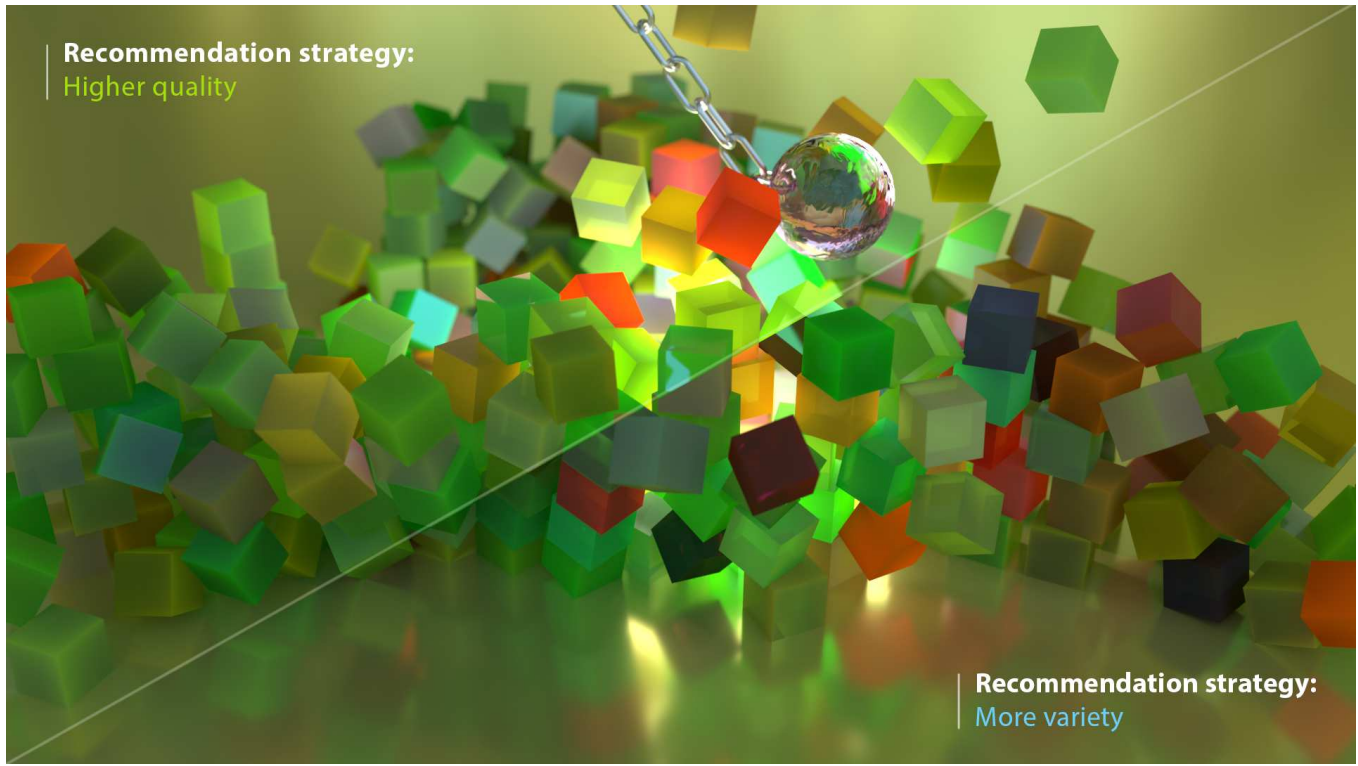

Fig. 3. This scene was generated using our automatic workflow. The recommendations of our system are controllable, i.e., the user can easily adjust the recommendation thresholds to fine-tune the amount of variety in the output distribution.

*Neural networks and rendering.* The recent resurgence of neural network-based learning techniques stimulated a large body of research works in photorealistic rendering. A class of techniques uses neural networks to approximate a selected aspect of light transport such as indirect illumination [Ren et al. 2013] or participating media [Kallweit et al. 2017]. To extend these endeavors, other works can be used to perform other related tasks, such as approximating sky models [Satỳlmỳs et al. 2017] or Monte Carlo noise filtering [Kalantari et al. 2015]. There is also a growing interest in replacing a greater feature set of the renderer with learning algorithms [Nalbach et al. 2017], which typically requires the presence of additional information, e.g., a number of auxiliary buffers. In our problem formulation, we are interested in a restricted version of this problem where geometry, lighting, and the camera setup are fixed and the BSDF parameters are subject to change. We show that in this case, it is possible to replace the entirety of the renderer without a noticeable loss of visual quality.

*GPR, GPLVM.* Gaussian Process Regression (GPR) is an effective learning method where prior knowledge of a problem can be harnessed via a covariance function, enabling high-quality regression using a modest amount of training samples. It offers useful solutions in the perimeter of computer graphics and machine learning – examples include performing super resolution [He and Siu 2011], analyzing and generating dynamical models for human motion [Wang et al. 2008], or synthesizing doodles and ocean waves [Anjyo and Lewis 2011]. Generative latent-space techniques proved to be highly useful in a variety of areas: they are able to design new fonts by using the *Gaussian Process Latent Variable Model* (GPLVM) to find low-dimensional structures in high-dimensional data and expose them to the user [Campbell and Kautz 2014], generate imaginary human faces and perform meaningful algebraic operations between them [Bojanowski et al. 2017], synthesize new shapes when given a database of examples [Averkiou et al. 2014], or suggest a selection of perceptually different parameter choices [Koyama et al. 2014; Marks et al. 1997]. It is clear that these methods are powerful tools in isolation – in this work, we show a novel combination of GPR, GPLVM and a Convolutional Neural Network that opens up the possibility of learning the material preferences of a user and offering a 2D latent space where the real-time fine-tuning of the recommended materials is possible.

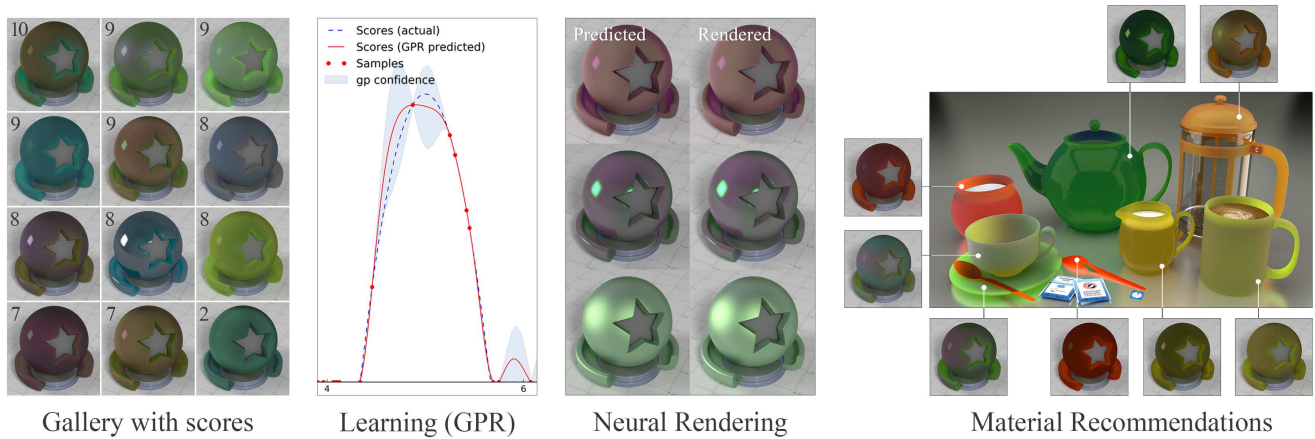

Fig. 4. In the first step, the user is presented with a gallery and scores the shown materials according to their taste. Then, a regression is performed to obtain a preference function via Gaussian Process Regression, which can be efficiently sampled for arbitrarily many new material recommendations. These recommendations can be visualized in real time using our neural network in a way that closely resembles the images rendered with global illumination. In the final step, the recommended materials can be conveniently assigned to an existing scene.

## 3 OVERVIEW

The overall workflow of our system consists of two stages (see also Fig. 4). In the *first stage*, the user is presented with a gallery and asked to assign scores to the shown materials. After choosing a threshold value to control the variability of the output, a set of recommendations are computed (Section 4.1) and visualized via neural rendering (Section 4.2). The recommendations depend entirely on the user scores and may span multiple material classes. If the user wishes to fine-tune a subset of the recommended materials, a latent space is inferred for low-dimensional exploration (Section 4.3). The user then evaluates the result:

(1) If the recommendations are acceptable – proceed to the next stage,
(2) If the recommendations need refinement – assign scores to the *newly proposed gallery* or *adjust past rankings* and compute a new round of recommendations.

In the *second stage*, the user can choose from two ways to assign the recommended materials to a scene:

(1) Automatic workflow – randomly assign the materials to the selected objects in the scene (Figures 1 (right) and 3). This is ideal for mass-scale material synthesis, when hundreds of materials are sought,
(2) Assisted workflow – assign the materials to the scene manually and perform fine-tuning via *variant generation* (Section 5.2), with color coding (Section 5.1). This is ideal when up to a few tens of materials are sought and strict control is required over the output (Figures 1 (left) and 9).

The final scene with the newly assigned material models is then to be rendered offline. In Section 4 we first present the three learning algorithms, which can be used for the automatic workflow by themselves, while in Section 5 we show how to combine them to provide an interactive system.

## 4 LEARNING ALGORITHMS FOR MATERIAL SYNTHESIS

In this section, we outline the three main pillars of our system: **Gaussian Process Regression** to perform material learning and recommendation, a **Convolutional Neural Network** variant for real-time image prediction, and the **Gaussian Process Latent Variable Model** to embed our high-dimensional shader inputs into a 2D latent space to enable the fine-tuning of a select set of recommended materials (Fig. 5). Table 1 summarizes the notation used throughout the paper.

### 4.1 Material Learning and Recommendation

In this section, we propose a combination of Gaussian Process Regression, Automatic Relevance Determination and Resilient Backpropagation to efficiently perform material learning and recommendation. We also show that these recommendations are easily controllable.

*Material learning.* Gaussian Process Regression (GPR) is a kernel-based Bayesian regression technique that leverages prior knowledge to perform high-quality regression from a low number of samples[3]. It can be used to approximate a preference function $u(\mathbf{x})$ from a discrete set of $n$ observations $\mathbf{U} = \left[ u(\mathbf{x}_1), u(\mathbf{x}_2), \ldots, u(\mathbf{x}_n) \right]^T$, each of which can be imagined as point samples of a Gaussian where $\mathbf{x}_i \in \mathbb{R}^m$ encode the parameters that yield a BSDF model. We created a parameter space similar to Disney's "principled shader" [Burley and Studios 2012] that comes in two versions: the $m = 19$ variant spans the most commonly used materials, i.e., a combination of diffuse, specular, glossy, transparent and translucent materials where the extended $m = 38$ version additionally supports procedurally textured albedos and displacements (see Section 6 and the supplementary materials). A Gaussian Process is given by its mean and covariance $k(\mathbf{x}, \mathbf{x}')$ that describes the relation between individual material samples $\mathbf{x}$ and $\mathbf{x}'$. A squared exponential covariance function is given as

$$k(\mathbf{x}, \mathbf{x}') = \sigma_f^2 \exp\left[-\frac{(\mathbf{x}-\mathbf{x}')^2}{2l^2}\right] + \beta^{-1}\delta_{xx'}, \quad (1)$$

with a given $\sigma_f^2$ variance and length scale $l$ where $\beta^{-1}\delta_{xx'}$ is an additional noise term enabled by the Kronecker delta to yield a positive definite covariance matrix. This means that a highly correlated pair $\|\mathbf{x}-\mathbf{x}'\| \approx 0$ yields the maximum of the function, i.e., $k(\mathbf{x}, \mathbf{x}') \approx \sigma_f^2 + \beta^{-1}\delta_{xx'}$, leading to a smooth function approximation as $u(\mathbf{x}) \approx u(\mathbf{x}')$. Conversely, if $\|\mathbf{x}-\mathbf{x}'\|$ is large, the two observations show negligible correlation, therefore for a rapidly decaying covariance function, $k(\mathbf{x}, \mathbf{x}') \approx 0$.

---
[3]Throughout this manuscript, we will use the terms *samples* and *observations* interchangeably.

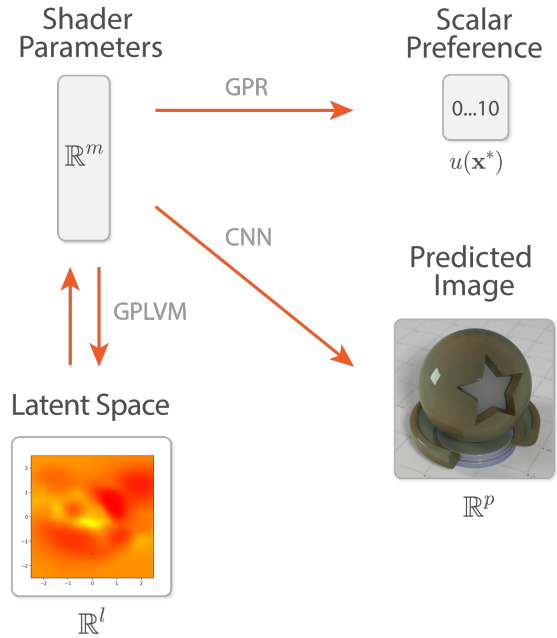

Fig. 5. A high-level overview of our pipeline: GPR is used to learn the user-specified material preferences and recommend new materials which are subsequently visualized using our Convolutional Neural Network. Optionally, GPLVM can be used to provide an intuitive 2D space for variant generation.

The covariance matrix $\mathbf{K}$ is given by all possible combinations of the point samples

$$\mathbf{K} = \begin{bmatrix} k(\mathbf{x}_1, \mathbf{x}_1) & k(\mathbf{x}_1, \mathbf{x}_2) & \dots & k(\mathbf{x}_1, \mathbf{x}_n) \\ k(\mathbf{x}_2, \mathbf{x}_1) & k(\mathbf{x}_2, \mathbf{x}_2) & \dots & k(\mathbf{x}_2, \mathbf{x}_n) \\ \vdots & \vdots & \ddots & \vdots \\ k(\mathbf{x}_n, \mathbf{x}_1) & k(\mathbf{x}_n, \mathbf{x}_2) & \dots & k(\mathbf{x}_n, \mathbf{x}_n) \end{bmatrix}, \quad (2)$$

the diagonal of $\mathbf{K}$ is therefore always $\sigma_f^2 + \beta^{-1}\delta_{xx'}$. The covariances for the unknown sample $\mathbf{x}^*$ are written as

$$\mathbf{k}_* = \left[k(\mathbf{x}^*, x_1), k(\mathbf{x}^*, x_2), \dots, k(\mathbf{x}^*, x_n)\right]^T, \quad (3)$$

| Symbol | Description | Type |
|---|---|---|
| $\mathbf{x}$ | BSDF description | Vector |
| $u^*(\mathbf{x})$ | Preference function (Ground truth) | Scalar |
| $u(\mathbf{x})$ | Preference function (GPR prediction) | Scalar |
| $n$ | Number of GPR samples | Scalar |
| $\mathbf{x}^*$ | Unknown BSDF test input | Vector |
| $\mathbf{U}$ | GPR training set | Matrix |
| $m$ | Input BSDF dimensionality | Scalar |
| $k(\mathbf{x}, \mathbf{x}')$ | Covariance function | Scalar |
| $\sigma_f^2$ | Variance | Scalar |
| $l$ | Length scale | Scalar |
| $\beta^{-1}$ | Noise term | Scalar |
| $\delta_{xx'}$ | Kronecker delta | Scalar |
| $\mathbf{K}$ | GPR covariance matrix | Matrix |
| $\theta$ | Covariance function parameterization | Vector |
| $\alpha$ | Learning rate | Vector |
| $\phi(\mathbf{x}^*)$ | CNN image prediction of a BSDF | Matrix |
| $\mathbf{X}$ | GPLVM training set | Matrix |
| $\mathbf{L}$ | Low dimensional latent descriptor | Matrix |
| $l$ | Latent space dimensionality | Scalar |
| $\mathcal{N}(x\|\mu, \sigma^2)$ | $\frac{1}{\sqrt{2\pi\sigma^2}}\exp(-\frac{(x-\mu)^2}{2\sigma^2})$ | Scalar |
| $z$ | Number of GPLVM samples | Scalar |
| $\mathbf{K}'$ | GPLVM covariance matrix | Matrix |
| $\psi(\mathbf{l}^*)$ | Mapping from latent to observed space | Vector |
| $m(\mathbf{x})$ | $\frac{1}{2}(u(\mathbf{x}) + u^*(\mathbf{x}))$ | Scalar |
| $\tau$ | Recommendation threshold | Scalar |
| $r$ | Grid resolution | Scalar |
| $s(\mathbf{x}^*, \mathbf{x}')$ | Similarity (CNN prediction) | Scalar |
| $u(\mathbf{x}')$ | Preference score (GPR prediction) | Scalar |

Table 1. Notation used throughout this paper, in order of occurrence.

(where $x_i \in \mathbf{x}^*$) and $k_{**} = k(\mathbf{x}^*, \mathbf{x}^*)$. We define a zero-mean Gaussian Process over $\left[\mathbf{U}, u(\mathbf{x}^*)\right]^T$ with the covariance function $k(\mathbf{x}, \mathbf{x}')$:

$$\begin{bmatrix} \mathbf{U} \\ u(\mathbf{x}^*) \end{bmatrix} \sim \mathcal{N}\left(0, \begin{bmatrix} \mathbf{K} & \mathbf{k}_*^T \\ \mathbf{k}_* & k_{**} \end{bmatrix}\right). \quad (4)$$

We seek $u(\mathbf{x}^*)$ leaning on the knowledge that the conditional probability $P(u(\mathbf{x}^*) | \mathbf{U})$ follows a Gaussian distribution, therefore the closed-form solution for $u(\mathbf{x}^*)$ and its variance is obtained by

$$u(\mathbf{x}^*) = \mathbf{k}_*^T \mathbf{K}^{-1} \mathbf{U},$$
$$\sigma(u(\mathbf{x}^*)) = k_{**} - \mathbf{k}_* \mathbf{K}^{-1} \mathbf{k}_*^T. \quad (5)$$

The quality of the regression depends on the choice and the parameterization of the covariance function. If $\theta = \{\sigma_f^2, l\}$ in (1) is chosen poorly, the result will suffer from severe over- or underfitting. To avoid this, a model selection step is performed to maximize the log-likelihood of the observed samples by choosing the appropriate hyperparameters, i.e.,

$$\log P(\mathbf{U}|\mathbf{x}, \boldsymbol{\theta}) = -\frac{1}{2}\mathbf{U}^T\mathbf{K}^{-1}\mathbf{U} - \frac{1}{2}\log|\mathbf{K}| - \frac{n}{2}\log 2\pi. \qquad (6)$$

The derivatives of the log-likelihood with respect to the hyperparameters and a learning rate $\boldsymbol{\alpha}$ are given by

$$\frac{\partial}{\partial \theta_j}\log P(\mathbf{U}|\mathbf{x}, \boldsymbol{\theta}) = -\frac{1}{2}\operatorname{tr}\left\{\left(\boldsymbol{\alpha}\boldsymbol{\alpha}^T\mathbf{K}^{-1}\right)\frac{\partial \mathbf{K}}{\partial \theta_j}\right\}, \qquad (7)$$

which can be used with gradient-based optimization techniques. In some cases, the user is looking for a class of material models where, for instance, the surface reflectance properties are of great importance and the choice of albedos is irrelevant (e.g., carpaint material variants). In this case, using one fixed length scale for all features leads to poor predictions. To this end, we have also used Automatic Relevance Determination [MacKay 1996; Neal 2012] and assigned a $\boldsymbol{\theta}$ to each dimension (with appropriate modifications to (1)) to open up the possibility of discarding potentially irrelevant features. However, this substantially increases the dimensionality of the optimization problem, to a point where classical methods such as L-BFGS-B [Byrd et al. 1995] and the Scaled Conjugate Gradient method [Møller 1993] prove to be ineffective: lower learning rates are theoretically able to find the desired minima, but slow down considerably in shallow regions while larger learning rates introduce oscillation near the minima. To this end, instead of using a fixed step size that is proportional to the local gradient, we adapt the learning rate based on the topology of the error function by using Resilient Backpropagation [Blum and Riedmiller 2013; Riedmiller and Braun 1992], a technique originally designed for training neural networks. This significantly reduces the number of required learning steps and has always succeeded finding usable minima in our experiments.

*Material recommendation.* Given enough learning samples, $u$ will resemble the true user preferences, therefore high-scoring material recommendations can be obtained by simply rejection-sampling it against a minimum acceptable score threshold. The rejection rates depend on the properties of $u$: choosing a high threshold will result in high-quality recommendations at the cost of higher rejection rates and decreased variety. Conversely, a larger variety of recommendations can be enforced by lowering the target threshold. Upon encountering a rejection ratio that is unacceptably high, we extend the rejection sampler with a low number of stochastic hillclimbing steps. This offers a considerably improved ratio at the cost of a minimal increase in sample correlation. Generally, we have found setting the recommendation threshold between 40% and 70% of the maximum possible score to be a good tradeoff between sample quality and variety (Fig. 3, higher quality (left) – 70%, more variety (right) – 40%).

### 4.2 Neural Networks and Rendering

After the learning and recommendation step have taken place, a gallery is generated with a prescribed amount of recommendations. The GPR and the recommendation steps take only a few seconds (Table 3), however, rendering all 300 recommendations (a typical number throughout our experiments) would take over 4 hours, which is a prohibitively long time for practical use. To cut down the time between the two steps, we introduce a neural network-based solution. Deep neural networks are universal function approximators that have proven to be remarkably useful for classification and regression problems. Typically, the dimensionality of the input is orders of magnitude larger than that of the output (e.g., image classification). Convolutional Neural Networks [LeCun et al. 1998] (CNNs) excel at solving problems of this form; their advantages include augmenting neurons with a receptive field to take advantage of the locality of information in images and their pooling operations that reduce the number of parameters in each layer. In this work, we are concerned with an atypical adjoint problem where images are to be predicted pixel by pixel from the shader input, $\phi \colon \mathbb{R}^m \to \mathbb{R}^p$, where the input dimensionality $m$ is in the order of tens, which is to be mapped to an output of $p$ dimensions, which represents the largest possible output that fits into the GPU memory along with its decoder network, i.e., in our case, a $410^2$ image with three color channels. We will refer to problems of this kind as *neural rendering*.

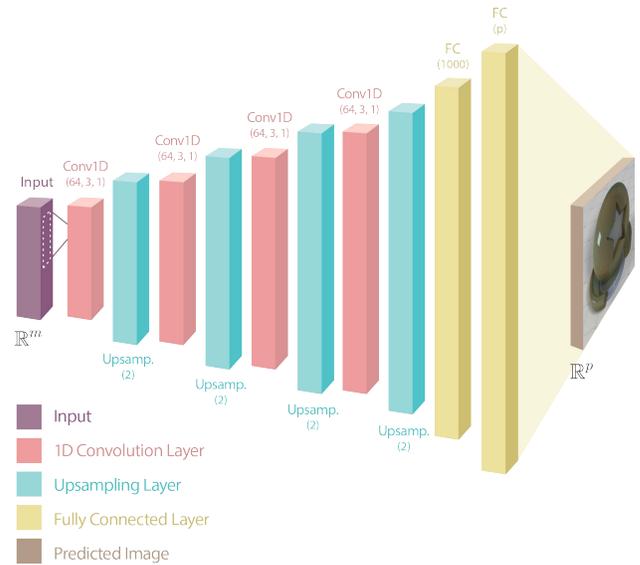

Fig. 6. Our decoder network takes the shader description as an input and predicts its appearance. Due to the fact that we have an atypical problem where the input shader dimensionality is orders of magnitude smaller than the output, the input signal is subjected to a series of 1D convolution and upsampling steps.

Neural rendering remains an unsolved problem in general. However, as our case is constrained to material modeling, the geometry and lighting setups can be kept constant, therefore it is possible to create a sufficiently deep network and training set to predict images that are nearly indistinguishable from the ones rendered until convergence with full global illumination. Using deconvolutional layers [Noh et al. 2015; Zeiler et al. 2010] would be a natural choice for $\phi$, however, as few of the most common software packages support it and unwanted artifacts may appear during image generation [Odena et al. 2016], instead, the input signal is subjected to a series of 1D convolutions, each followed by an upsampling layer to inflate

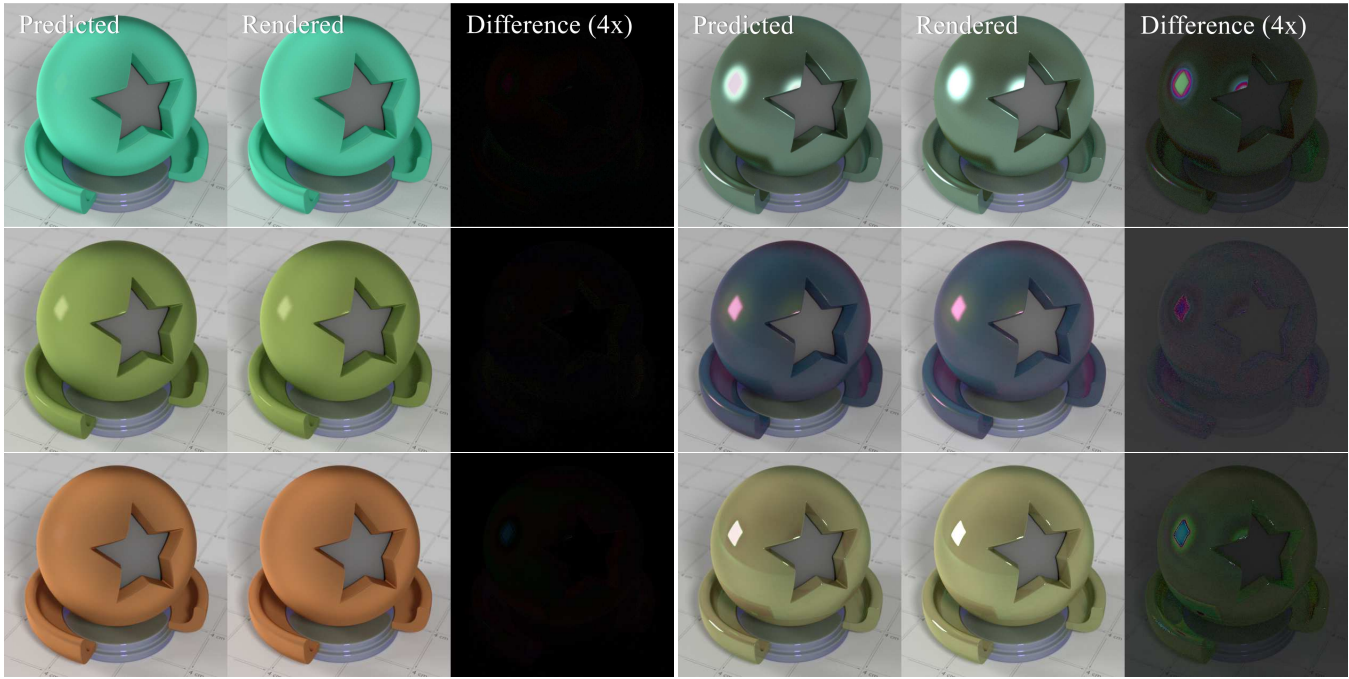

Fig. 7. The best (left side) and worst-case (right side) predictions by our neural network on a set of 250 images. Mean PSNR: 37.96dB, minimum: 26.05dB, maximum: 48.70dB.

the number of parameters as we advance deeper into the network (Fig. 6).

This architecture is similar to the decoder part of Convolutional Autoencoders [Masci et al. 2011] and can be described with the shorthand notation of $4x\{Conv1D(64, 3, 1) - Upsampling(2)\}$ – $FC(1000)$ – $FC(410^2 \cdot 3)$, where the parameters of the convolutional layer are *number of filters*, *spatial kernel size* and *strides*, respectively. These layers use exponential linear units [Clevert et al. 2015] with Glorot-initialization [Glorot and Bengio 2010] and are trained via the Adam optimizer [Kingma and Ba 2014] ($lr=10^{-3}$, $\beta_1=0.9$, $\beta_2=0.999$, $\epsilon=10^{-8}$, decay=0). Normally, a network of this size introduces severe overfitting, even in the presence of $\mathcal{L}_1/\mathcal{L}_2$ regularization [Nowlan and Hinton 1992; Zou and Hastie 2005] or dropout [Srivastava et al. 2014], especially if learning takes place on a given, fixed dataset. However, as we can create our own dataset, i.e., a potentially infinite number of shader-image pairs via rendering, we are able to evade this problem by creating a sufficiently large training set. In the interest of efficiency, we have generated 45000 LDR shader-image pairs with a spatial resolution of $410^2$ and 250 samples per pixel over 4 weeks on a consumer system with a NVIDIA GeForce GTX TITAN X GPU, and since no means were required to prevent overfitting, our training process converged to $10^{-2}$ (RMSE), a negligible but non-zero $\mathcal{L}_2$ training and validation loss (where a zero loss would mean containing all the noise from the training set) in less than 30 hours. Our validation set contained 2500 images and the measured losses correlated well with real-world performance. This pre-trained network can be reused as long as the shader description remains unchanged and the inference of a new image typically takes 3 to 4 milliseconds. A further advantage of this architecture is that similarly to Denoising Autoencoders [Vincent et al. 2010], it also performs *denoising* on the training samples.

A key observation is that the more layers the neural network contains, the more high-frequency details it will be able to represent (a similar effect has been observed in Fig 4., Saito et al. [2016]). This means that our proposed neural network contains enough layers to capture the important features to maximize visual quality, but not enough to learn the high-frequency noise contained in the dataset. Normally, this denoising process requires the presence of auxiliary feature buffers and longer computation times [Sen and Darabi 2012] or other deep learning approaches using more complex architectures [Bako et al. 2017]. In our case, this step does not require any additional complexity and is inherent to the structure of our network. Beyond producing less noisy images in real time, this is also a significant advantage in easing the computational load of creating new datasets as the training images do not have to be rendered until convergence. We took advantage of this by using only 250 samples per pixel for each training image, which took six times less than the standard 1500 samples that would be required for perfectly converged training samples, cutting down the total time to produce the training set from 24 to 4 weeks.

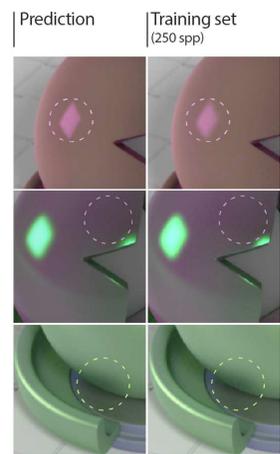

### 4.3 Latent Space Variant Generation

After being presented with a gallery of material models, the user may find that some recommendations are close to their preference, but some require fine-tuning to fit their artistic vision for a scene. Adjusting the materials by hand requires domain expertise and a non-trivial amount of trial and error. In our solution, we seek a dimensionality reduction technique that maps the shader inputs into a 2D latent space where similar materials can be intuitively explored. Non-linear manifold learning techniques come as a natural choice for this kind of problem, however, the most well-known methods [Belkin and Niyogi 2003; Maaten and Hinton 2008; Tenenbaum et al. 2000] could not find coherent structures in 2D. Beyond that, these techniques are unable to interpolate between the embedded samples, therefore they would only be useful for visualization in a latent space, but not for exploration. Dimensionality reduction with autoencoders [Hinton and Salakhutdinov 2006] would require orders of magnitude more samples to work properly (we have at most a few tens at our disposal), which is infeasible as it would be too laborious for the user to provide that many scores through the presented gallery.

The Gaussian Process Latent Variable Model (GPLVM) [Lawrence 2004] is a non-linear dimensionality reduction technique which is able to embed a few tens of high-scoring materials from the gallery $\mathbf{X} = \begin{bmatrix} \cdots \mathbf{x}_i \cdots \end{bmatrix}^T$ with $\mathbf{x}_i \in \mathbb{R}^m$ into a set of low dimensional latent descriptors $\mathbf{L} = \begin{bmatrix} \cdots \mathbf{l}_i \cdots \end{bmatrix}^T$ with $\mathbf{l}_i \in \mathbb{R}^l$. Typically, $m \gg l$, in our case, $m = 19$ and $l = 2$ to make sure that variant generation can take place conveniently on a 2D plane. The likelihood of the high-dimensional data using $z$ high-scoring training samples is given as

$$P(\mathbf{X}|\mathbf{L}, \theta) = \prod_{i=1}^{z} \mathcal{N}\left(\mathbf{X}_i \mid \mathbf{0}, \ \mathbf{K}' + \beta^{-1}\mathbf{I}\right), \tag{8}$$

where $\mathbf{K}'$ is a covariance matrix similar to $\mathbf{K}$ containing elements akin to (1) with a substitution of $k(\mathbf{x}, \mathbf{x}') \to k(\mathbf{l}, \mathbf{l}')$. In this case, the optimization takes place jointly over the latent values in $\mathbf{L}$ and $\theta$, i.e.,

$$\mathbf{L}^*, \theta^* = \underset{\mathbf{L}, \theta}{\operatorname{argmax}} \ \log\left[P(\mathbf{X}|\mathbf{L}, \theta)\right]. \tag{9}$$

Even though PCA disregards the non-linear behavior of BSDF parameters [Lafortune et al. 1997], it serves as a formidable initial guess for $\mathbf{L}$ [Lawrence 2004] and $\theta$ is initialized with a wide prior. Beyond the ability to learn efficiently from a handful a samples, a further advantage of this method is that a new mapping can be made between the latent and observed space $\mathbf{x}^* = \psi(\mathbf{l}^*)$, i.e.,

$$\begin{bmatrix} \mathbf{X} \\ \mathbf{x}^* \end{bmatrix} \sim \mathcal{N}\left(0, \begin{bmatrix} \mathbf{K}' & \mathbf{k}'^T_* \\ \mathbf{k}'_* & \mathbf{k}'_{**} \end{bmatrix}\right), \tag{10}$$

yielding the final closed-form solution

$$\psi(\mathbf{l}^*) = \mathbf{k}'^T_* \mathbf{K}'^{-1} \mathbf{X},$$
$$\sigma(\psi(\mathbf{l}^*)) = \mathbf{k}'_{**} - \mathbf{k}'_* \mathbf{K}'^{-1} \mathbf{k}'^T_*. \tag{11}$$

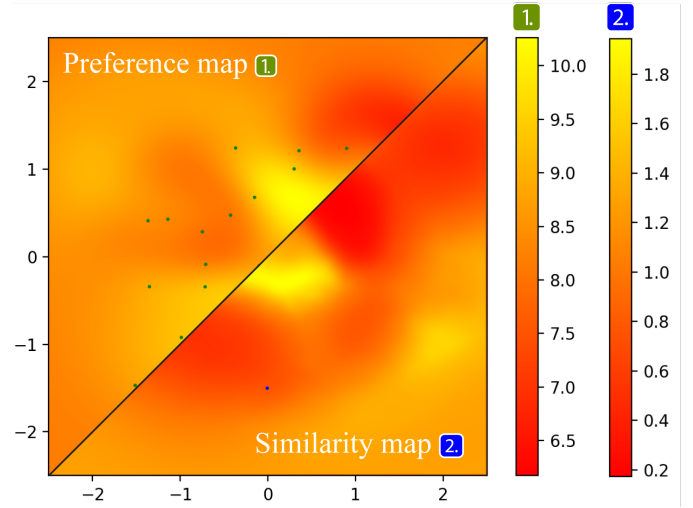

Fig. 8. Endowing the latent space with the expected preferences (upper left) and similarities (lower right). The green dots represent the embedded training samples, where the blue dot shows the reference input material to be fine-tuned.

By storing and reusing $\mathbf{K}'^{-1}$, this mapping can be done in negligible time, which allows the user to rapidly generate new material variants in this 2D latent space.

## 5 INTERACTIVE LATENT SPACE EXPLORATION

We have introduced a system using three learning algorithms in isolation. GPR enables material learning and recommendation from a few learning samples, the CNN opens up the possibility of neural rendering, and the high-dimensional shader can be non-linearly embedded in a 2D latent space using GPLVM. In this section, we propose novel ways to combine these algorithms to obtain a system for rapid mass-scale material synthesis and variant generation. A more rigorous description of the final algorithm is presented in Appendix A.

### 5.1 GPLVM Color Coding

When using GPLVM, each point in the 2D latent space corresponds to an $m$-dimensional vector that describes a material model. Since we have used GPR to learn the correspondence between these materials and user preferences, it is possible to combine these two techniques to obtain the expected scores for these samples. This combination enables a useful visualization of the latent space where these expected preferences appear in the form of color coding. This *preference coding* is useful to highlight regions of the latent space that encode favorable materials, however, when fine-tuning a chosen material, the requirement of obtaining *similar* materials is of equal importance. Since our CNN is able to predict images in real time, we propose subdividing the latent space into a 2D grid, where an image can be predicted in each gridpoint. This image can be compared to the image of the material we wish to fine-tune via a distance metric of choice (e.g., $\mathcal{L}_1/\mathcal{L}_2$), therefore, the latent space can thus be endowed with *similarity information* as well.

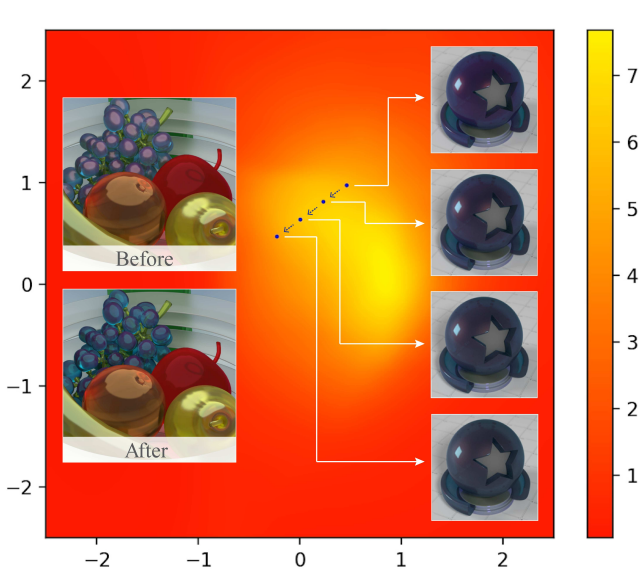

Fig. 9. The color-coded 2D latent space can be explored in real time by the user for variant generation. The vividness of the recommended grape material can be fine-tuned rapidly without any domain knowledge.

The preference map is global, i.e., it remains the same regardless of our input material where the similarity map depends on our current material that is used as a starting point (Fig. 8). The product of these two maps offers an effective way to create variants of a source material that are similar, and are highly preferred according to the learned preferences.

## 5.2 Real-Time Variant Generation

Exploring the 2D latent space of the learned materials is of limited usefulness when the user has to wait for 40-60 seconds for each new image to be rendered. A typical use case involves sweeping motions that require near instantaneous feedback from the program. As the 2D output of the GPLVM can be projected back to the high-dimensional material space (with a dimensionality increase of $l \rightarrow m$), this output can be combined with our CNN, which provides real-time image predictions to allow efficient exploration in the latent space with immediate feedback. We demonstrate several such workflows in our supplementary video.

## 6 RESULTS

In this section, we evaluate the three learning algorithms in isolation and demonstrate the utility of our whole system by recording modeling timings against the classical workflow for three practical scenarios. Furthermore, we also discuss how the proposed system handles our extended shader.

*Learning material spaces.* After having obtained $u$, we are interested in measuring the quality of the regression by relating it to the true user preference function $u^*$. By normalizing both functions and treating them as probability distributions, the Jensen-Shannon divergence (JSD) yields a suitable metric to distinguish how much information is lost if $u$ is used as a proxy for the unknown $u^*$, i.e.,

$$\mathrm{JSD}\bigl(u(\mathbf{x}) \,\|\, u^*(\mathbf{x})\bigr) = $$
$$\frac{1}{2}\int_{-\infty}^{+\infty} u(\mathbf{x}) \log \frac{u(\mathbf{x})}{m(\mathbf{x})}\, d\mathbf{x} + \frac{1}{2}\int_{-\infty}^{+\infty} u^*(\mathbf{x}) \log \frac{u^*(\mathbf{x})}{m(\mathbf{x})}\, d\mathbf{x}, \quad (12)$$

where $m(\mathbf{x}) = \frac{1}{2}\bigl(u(\mathbf{x}) + u^*(\mathbf{x})\bigr)$. We have recorded the JSD produced by minimizing (7) with RProp, L-BFGS-B and the Scaled Conjugate Gradient method and found that all three techniques are competitive for the lower-dimensional case, i.e., $m = 19$. In the case of the extended shader, RProp consistently outperformed L-BFGS-B and SCG, both of which often got stuck in poor local minima even when being rerun from many randomized initial guesses (Table 2). For the high-dimensional cases with over 200 training samples, SCG did not always converge despite a non-singular $K$ due to round-off errors. We have used two challenging cases to demonstrate the utility of our system by learning the material space of glassy and translucent materials. These cases are considered challenging in a sense that these materials are relatively unlikely to appear via random sampling: in the glassy use case, 81% of the samples in the initial gallery were scored zero. This ratio was 90% for the translucent case. We have scored 1000 glassy and translucent materials on a scale of 0 to 10 to use as a ground truth dataset, where the first 250 samples were used as training data for the GPR. In each case, our technique was able to generate high-quality recommendations from 46 (glassy) and 23 (translucent) non-zero observations. The remaining 750 samples were used for cross-validation to compute a reliable estimate of the JSD. In both cases, the training took 7.22s and as a result, an arbitrarily large gallery of recommendations can be generated in 0.06s per recommendation on a mid-range consumer Intel Core i5-6600 CPU (see Table 3 for a detailed breakdown). The metals and minerals scene in Fig. 12 showcases a *multi-round* learning scenario where the recommendation gallery was scored and re-used to generate a second, more relevant gallery of materials (all other cases use one round of scores). The images and scores used in the training sets are available in the supplementary materials.

| Scene | m | n | RProp | L-BFGS-B | SCG |
|---|---|---|---|---|---|
| Glassy | 19 | 150 | **0.08** | 0.10 | **0.08** |
| Glassy | 19 | 250 | 0.09 | 0.09 | **0.08** |
| Glassy | 19 | 500 | 0.07 | 0.07 | 0.07 |
| Translucent | 19 | 150 | **0.17** | 0.18 | 0.18 |
| Translucent | 19 | 250 | **0.19** | 0.19 | – |
| Translucent | 19 | 500 | 0.17 | 0.17 | 0.17 |
| Glassy | 38 | 150 | **0.41** | 0.58 | 0.58 |
| Glassy | 38 | 250 | **0.35** | 0.57 | 0.57 |
| Glassy | 38 | 500 | **0.14** | 0.53 | 0.38 |
| Metals/Minerals | 38 | 150 | **0.53** | 0.55 | 0.55 |
| Metals/Minerals | 38 | 250 | **0.44** | 0.52 | – |
| Metals/Minerals | 38 | 500 | **0.32** | 0.62 | 0.60 |

Table 2. All three optimization techniques produce competitive JSD values in the lower dimensional case (i.e., $m = 19$). In the case of the extended shader ($m = 38$), RProp consistently outperforms L-BFGS-B and SCG regardless of the number of training samples ($n$).

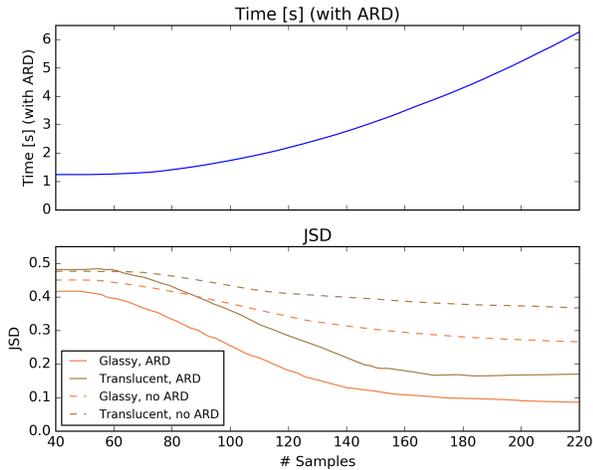

Fig. 10. Even for more challenging cases, the presence of Automatic Relevance Determination stabilizes the GPR reconstruction quality around 150-250 training samples.

*Variant generation.* A synthesized material of choice can be fine-tuned via variant generation in our 2D latent space. The exploration is guided by two different kinds of color-coding and real-time previews of the final materials. We demonstrate the usefulness of this element of our system in a real-world scenario by reducing the vividness of the grape material in the glassy still life scene (Fig. 9 and supplementary video). The color-coded regions denote outputs that are preferred *and* similar to the input material and form an island that is easy to explore. The preference and similarity maps are computed on a $50^2$ and $20^2$ grid respectively using an NVIDIA GeForce GTX TITAN X GPU and are subjected to bilinear interpolation in our visualizations.

*Neural rendering comparisons.* Because of our restricted problem definition, our neural network can mimic a global illumination renderer with high-quality predictions, and does not require retraining when combined with a sufficiently expressive principled shader. Three average case predictions are shown in Fig. 4 (middle). Three of the best- and worst-case predictions and their difference images as well as PSNR values are reported on a set of 250 images in Fig. 7. Querying this neural network takes 3-4ms on average and in every case, the predicted images were close to indistinguishable from the ground truth. An additional advantage our proposed architecture is that even though the training set contained moderately noisy images (250 spp, visible in the floating image in Section 4.2 and zooming in to the "Gallery with scores" part of Fig. 4), the predicted images appear smoother.

*Modeling and execution time.* To show that our system is useful for novice and expert users alike, we recorded the time required to model 1, 10, and 100 similar materials using Disney's "principled" shader [Burley and Studios 2012] against our technique. The two main user types to be compared to is a novice user who has no knowledge of light transport and material modeling, and an expert with significant experience in material modeling. Both were allowed several minutes to experiment with the principled shader before starting. The novice and expert users took 161s and 52s to obtain one prescribed base material model (a slightly scattering blue glass material with a small, non-zero roughness). Creating subsequent variants of this material took 29s and 19s where most of the time was spent waiting for a reasonably converged rendered image to show the minute differences between the base material and the new variant. Using our technique, in the presented gallery, it took an average of 2s to score a non-zero sample and 0.4s for a sample with the score zero. Typically, in our workflows, 250 observations were used to learn the material spaces and provide recommendations therefore we based our timings on that number. When only one material model is sought, novice users experience roughly equivalent modeling times when using our proposed system. In the case of mass-scale material synthesis, modeling times with our system outperform expert users. Beyond cutting down the time spent with material modeling, our system provides several other advantages over the traditional workflow: it does not require any domain expertise, provides real-time denoised previews throughout the process, and during scoring, the users are exposed to a wider variety of examples. This last advantage is especially useful for novices who do not necessarily have a prior artistic vision and are looking for inspiration. We also note that the workflow timings are often even more favorable as 150 samples are enough to provide satisfactory results for learning challenging material spaces (Fig. 10). Our intention in Fig. 11 was to show that our timings are appealing even in the more pessimistic cases.

| Stage | Time [$s$] | Size |
|---|---|---|
| GPR | 7.22 | 250 |
| Recommendation | 0.06 / 17.4 | 1 / 300 |
| GPLVM | 1.96 | 16 |
| CNN | 0.04 | $410^2$ |
| Preference coding | 2.75 | $50^2$ |
| Similarity coding | 8.15 | $20^2$ |
| Sum | 20.18 / 37.52 | |

Table 3. Execution times for different stages of the proposed method. The 'Size' column stands for the size of the problem at hand, i.e., training samples for GPR and GPLVM, number of recommended materials, image resolution for the CNN, and 2D spatial resolutions for preference and similarity coding.

*Adding displacements.* In the results shown so far, we have used a shader with $m = 19$ as a basis for the training process, however, the GPR and GPLVM steps are capable of learning significantly higher-dimensional inputs. To demonstrate this, we have created a higher-dimensional SVBRDF shader that includes procedural textures and displacements. This shader ($m = 38$) is even more in line with our design principles, i.e., more expressive at the cost of being less intuitive, which is alleviated by using learning methods instead of interacting with it directly. In this case, material recommendation and learning steps still perform well – we have used 500 samples to learn metals and minerals and 150 samples for glittery Christmas ornament materials (Fig. 12). The limitation of this extended shader is the fact that it is ample in localized high-frequency details that our CNN was unable to represent. We note that this is a hardware limitation, and by adding more layers, more of these details are expected to appear, our architecture will therefore be able to predict these features as further hardware advancements take place, noting that this may require a higher sampling rate for the training set. After the recommendation and material assignment steps, displacements can be easily added by hand to the simpler shader setup (as shown in the supplementary video).

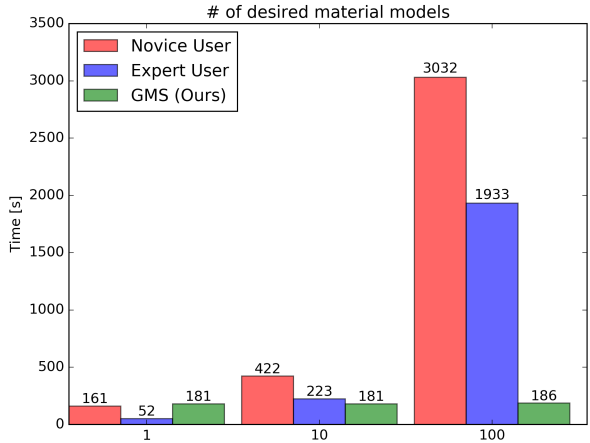

Fig. 11. Time taken to generate 1, 10, and a 100 similar materials by hand for users of different experience levels versus our technique (with the GPR and recommendations steps).

## 7 FUTURE WORK

Our technique starts out by showing a randomly generated gallery to the user to obtain a set of scores. These BSDFs are sampled with uniform distribution. We have experimented with improving it with an active learning scheme [Kapoor et al. 2007] to introduce adaptivity to the sampling process, making the newer gallery elements more relevant as they depend on the user-specified scores. For instance, the sampler can be equipped with novelty search [Lehman and Stanley 2008, 2011] to aggressively look for unexplored regions that may be preferred by the user. Furthermore, after obtaining the first few samples, quick GPR runs can take place, and instead of standard uniform distribution, the upcoming gallery images can be drawn from this learned intermediate distribution. We have found this scheme highly effective, and in the supplementary material, we provide a case with a "two-round" recommendation run for metals and minerals. Novice users can be further aided through automatic variant generation of new material(s) $\mathbf{x}'$ to fine-tune an input $\mathbf{x}^*$ by maximizing $\lambda s(\mathbf{x}^*, \mathbf{x}') + (1 - \lambda) u(\mathbf{x}')$ for a multitude of different $\lambda \in (0, 1)$ choices instead of relying solely on maximizing $u(\mathbf{x}')$ for recommendations (see Table 1 for details on the notation). By using GPR and GPLVM, not only the regressed outputs, but also their confidence values can be visualized (Equations (5) and (11)), or used as additional information for active learning. High-dimensional measured BRDF representations may also be inserted into the system. Due to the increased parameter count, the GPR should be replaced by a regressor that scales more favorably with the number of input dimensions, e.g., a deep neural network. As this also requires the presence of more training samples during the regression step, nearby regions in the similarity map could be channeled back to the neural network as additional data points. Since there is no theoretical resolution limit for the image predictions, our CNN can be retrained for higher resolutions as GPU technology improves, leaving room for exciting future improvements. To further enhance the quality of the neural network outputs, we have implemented the "late fusion" model in Karpathy et al.'s two-stream architecture [2014] and experienced measurable, but marginal improvements. As this area is subject to a significant volume of followup works, we expect that this direction, alongside with rapid improvements in GPU technology will lead to the possibility of predicting outputs with more high-frequency details in full HD resolution in the near future. Variable light source types, positions, and camera angles can also be learned by the neural network to enhance the quality of gallery samples by showing animations instead of stationary images.

## 8 CONCLUSIONS

We have proposed a system for mass-scale material synthesis that is able to rapidly recommend new material models after learning the user preferences from a modest number of samples. Beyond this pipeline, we also explored combinations of the three used learning algorithms, thereby opening up the possibility of real-time material visualization, exploration and fine-tuning in a 2D latent space. Furthermore, the system works with arbitrary BSDF models and is future-proof, i.e., preference learning and recommendation works with procedural textures and displacements, where the resolution and visualization quality is expected to further improve as the graphics card compute power and on-board VRAM capacities grows over time. Throughout the scoring and recommendation steps, the users are shown noise-free images in real time and the output recommendation distribution can be controlled by a simple change of a parameter. We believe this feature set offers a useful solution for rapid mass-scale material synthesis for novice and expert users alike and hope to see more exploratory works harnessing and combining the advantages of multiple learning algorithms in the future.

## ACKNOWLEDGMENTS

We would like to thank Robin Marin for the material test scene and Vlad Miller for his help with geometry modeling, Felícia Zsolnai–Fehér for improving the design of many figures, Hiroyuki Sakai, Christian Freude, Johannes Unterguggenberger, Pranav Shyam and Minh Dang for their useful comments, and Silvana Podaras for her help with a previous version of this work. We also thank NVIDIA for providing the GPU used to train our neural networks. This work was partially funded by Austrian Science Fund (FWF), project number P27974. Scene and geometry credits: Gold Bars – JohnsonMartin, Christmas Ornaments – oenvoyage, Banana – sgamusse, Bowl – metalix, Grapes – PickleJones, Glass Fruits – BobReed64, Ice cream – b2przemo, Vases – Technausea, Break Time – Jay–Artist, Wrecking

Ball – floydkids, Italian Still Life – aXel, Microplanet – marekv, Microplanet vegetation – macio.

## A GMS PSEUDOCODE

To maximize reproducibility, we provide the pseudocode of our system below. Note that the symbol "_" in lines 7, 14 and 16 refer to throwing away part of the function return value (i.e., using only the first dimension of the output of a 2D function).

**Algorithm 1** Gaussian Material Synthesis

1: **given** $\tau, r$ ▷ Recommendation threshold, grid resolution
2: **for** $i \leftarrow 1$ to $k$ **do** ▷ $k$ GPR training samples
3:    Generate random BSDF $\mathbf{x}$
4:    $\mathbf{U}_i \leftarrow u(\mathbf{x})$
5: $\mathbf{X} \leftarrow \{\mathbf{x} \in \mathbf{U} \mid \mathbf{x} > \tau\}$ ▷ GPLVM training set
6: **for** $i \leftarrow 1$ to $l$ **do** ▷ Recommendations
7:    **while** SCORE($\mathbf{X}, \mathbf{x}, \beta^{-1}$) $< \tau, \_$ **do**
8:      Generate random $\mathbf{x}$
9:    $\mathbf{R}_l \leftarrow \mathbf{x}$
10: Choose $\mathbf{x}^* \in \mathbf{R}$ for variant generation
11: Display $\phi(\mathbf{x}^*)$
12: **for** $i, j \leftarrow 1$ to $r$ **do** ▷ GPLVM color coding, Section 5.1
13:    **init** $\mathbf{G}_{ij}$ gridpoint with coordinates $i, j$ and resolution $r$
14:    $\mathbf{x}', \_ \leftarrow$ LATENT($\mathbf{X}, \mathbf{G}_{ij}, 2, \beta^{-1}$)
15:    $s(\mathbf{x}^*, \mathbf{x}') \leftarrow ||\phi(\mathbf{x}^*) - \phi(\mathbf{x}')||_{\mathcal{L}_2}$ ▷ Similarity (CNN)
16:    $u(\mathbf{x}'), \_ \leftarrow$ SCORE($\mathbf{U}, \mathbf{x}', \beta^{-1}$) ▷ Preference (GPR)
17:    $T_{ij} \leftarrow s(\mathbf{x}^*, \mathbf{x}')u(\mathbf{x}')$ ▷ Store product coding
18:    **assign** color $\mathbf{T}_{ij}$ to gridpoint $\mathbf{G}_{ij}$
19: **while** Given user displacement $\boldsymbol{\delta}$ **do** ▷ Explore latent space
20:    Display $\phi(\mathbf{x}^* + \boldsymbol{\delta})$

**Algorithm 2** Scoring a new BSDF

1: **function** SCORE($\mathbf{U}, \mathbf{x}^*, \beta^{-1}$) ▷ Score new BSDF, Section 4.1
2:    **init** $\mathbf{K}, \theta$
3:    $\theta^* \leftarrow \arg\max_\theta \log \left[ P(\mathbf{U} | \mathbf{x}, \theta) \right]$
4:    $u(\mathbf{x}^*) \leftarrow \mathbf{k}_*^T \mathbf{K}^{-1} \mathbf{U}$
5:    $\sigma(u(\mathbf{x}^*)) \leftarrow k_{**} - \mathbf{k}_* \mathbf{K}^{-1} \mathbf{k}_*^T$
6:    **return** $u(\mathbf{x}^*), \sigma(u(\mathbf{x}^*))$

**Algorithm 3** Latent space mapping

1: **function** LATENT($\mathbf{X}, \mathbf{x}^*, l, \beta^{-1}$) ▷ Latent mapping, Section 4.3
2:    **given** $\mathbf{x}^* = \psi(\mathbf{l}^*)$
3:    **init** $\mathbf{K}', \theta$
4:    $\mathbf{L}^*, \theta^* \leftarrow \arg\max_{\mathbf{L}, \theta} \log \left[ P(\mathbf{X} | \mathbf{L}, \theta) \right]$
5:    $\psi(\mathbf{l}^*) \leftarrow \mathbf{k}_*'^T \mathbf{K}'^{-1} \mathbf{X}$
6:    $\sigma(\psi(\mathbf{l}^*)) \leftarrow k_{**}' - \mathbf{k}_*' \mathbf{K}'^{-1} \mathbf{k}_*'^T$
7:    **return** $\psi(\mathbf{l}^*), \sigma(\psi(\mathbf{l}^*))$

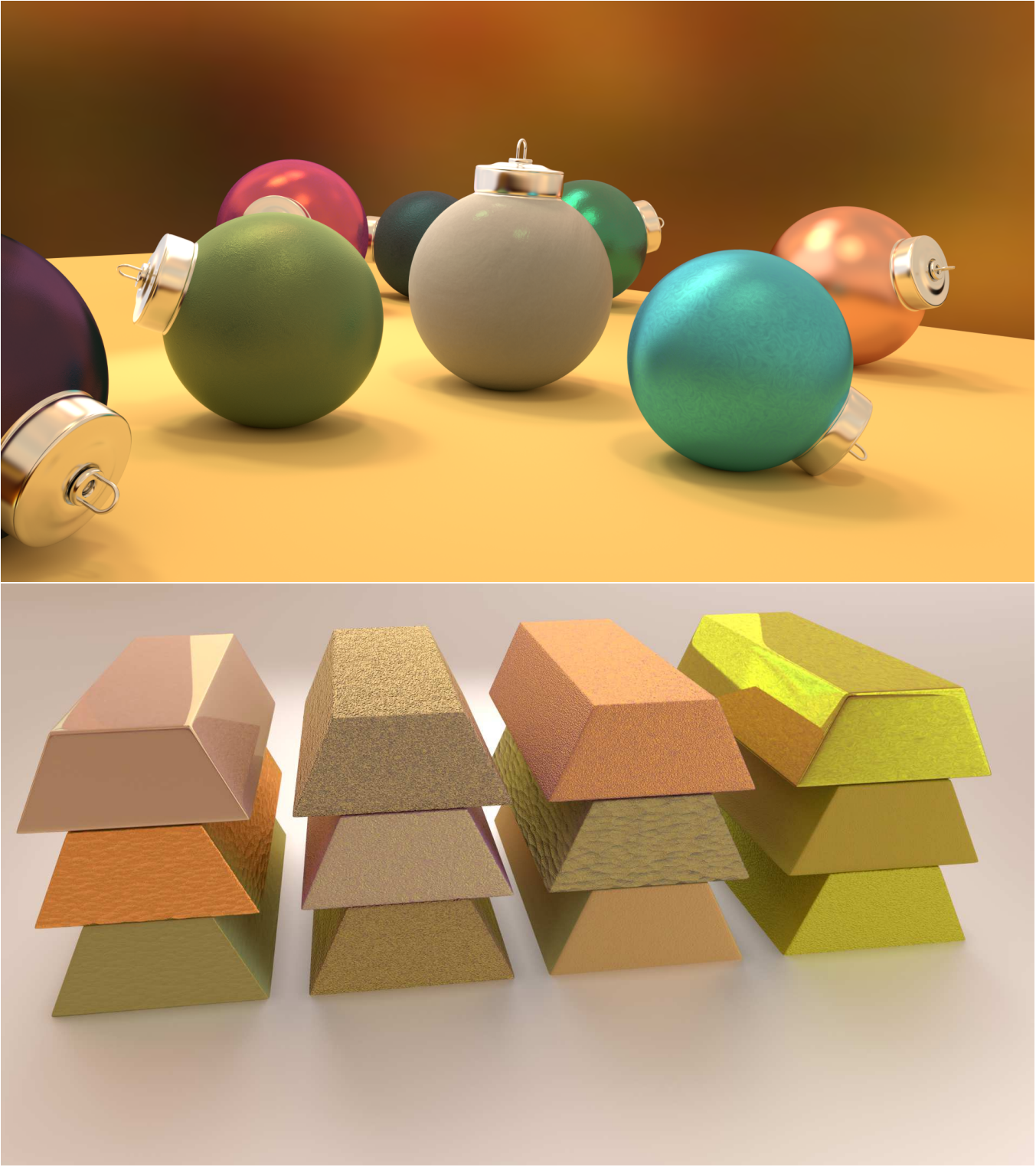

Fig. 12. Synthesized glittery materials (above) followed by metals and minerals (below) using our extended shader.